\pdfoutput=1

\documentclass[11pt]{article}

\usepackage[final]{acl}
\usepackage{amssymb}
\usepackage{amsmath}
\usepackage{amsthm}
\usepackage{algpseudocode}
\usepackage[ruled,vlined]{algorithm2e}

\usepackage{etoolbox} 

\makeatletter
\patchcmd{\algocf@makecaption@ruled}{\hsize}{\textwidth}{}{} 
\patchcmd{\@algocf@start}{-1.5em}{0em}{}{} 
\makeatother 
\RestyleAlgo{ruled}
\usepackage{hyperref}
\usepackage{times}
\usepackage{latexsym}
 \usepackage{multirow} 
\usepackage[T1]{fontenc}
\usepackage{comment}
\usepackage[utf8]{inputenc}

\usepackage{microtype}

\usepackage{inconsolata}

\usepackage{graphicx}

\title{Task-Informed Anti-Curriculum by Masking\\ 
Improves Downstream Performance on Text}

\author{Andrei Jarca, Florinel Alin Croitoru \and {Radu Tudor Ionescu}\thanks{Corresponding author: \texttt{raducu.ionescu@gmail.com}.}\\
  University of Bucharest\\
 Bucharest, Romania}

\begin{document}
\maketitle
\begin{abstract}
Masked language modeling has become a widely adopted unsupervised technique to pre-train large language models (LLMs). However, the process of selecting tokens for masking is random, and the percentage of masked tokens is typically fixed for the entire training process. In this paper, we propose to adjust the masking ratio and to decide which tokens to mask based on a novel task-informed anti-curriculum learning scheme. First, we harness task-specific knowledge about useful and harmful tokens in order to determine which tokens to mask. Second, we propose a cyclic decaying masking ratio, which corresponds to an anti-curriculum schedule (from hard to easy). We exemplify our novel task-informed anti-curriculum by masking (TIACBM) approach across three diverse downstream tasks: sentiment analysis, text classification by topic, and authorship attribution. Our findings suggest that TIACBM enhances the ability of the model to focus on key task-relevant features, contributing to statistically significant performance gains across tasks. We release our code at \url{https://github.com/JarcaAndrei/TIACBM}.
\end{abstract}

\setlength{\abovedisplayskip}{3.0pt}
\setlength{\belowdisplayskip}{3.0pt}

\section{Introduction}
\vspace{-0.1cm}
Nowadays, masked language modeling (MLM) \cite{Devlin-NAACL-2019} is one of the most popular frameworks used to pre-train language models, as it enables the use of vast amounts of unlabeled data. However, the process of selecting tokens for masking is generally based on random selection, while the percentage of masked tokens is typically fixed for the entire training process \cite{Wettig-ECAL-2023}. To the best of our knowledge, there are only two studies attempting to dynamically adapt the masking ratio \cite{Ankner-EACL-2024,Yang-ACL-2023}. These studies concur that the optimal schedule is to use a decaying masking ratio during training. Interestingly, we find that this observation is deeply connected to the curriculum learning paradigm. Curriculum learning is a training strategy formulated by \citet{Bengio-ICML-2009}, where neural models learn the data in a systematic manner, starting with easy samples and gradually adding more difficult samples as the learning progresses. Intuitively, using a higher masking ratio makes the learning task more difficult. Hence, employing a decaying masking ratio corresponds to an anti-curriculum strategy \cite{Liu-CVPR-2022, Soviany-IJCV-2022}.

In this paper, we further develop and explore anti-curriculum learning based on MLM to fine-tune pre-trained models on downstream tasks. We propose a novel task-informed anti-curriculum by masking (TIACBM) scheme, which employs a cyclic decaying masking ratio and relies on task-specific knowledge to decide which tokens to mask. Our most important contribution is to harness task-specific knowledge about useful and harmful tokens in order to select tokens for masking. For example, in sentiment analysis, the masking probability of a token is determined based on its polarity scores recorded in SentiWordNet 3.0 \cite{Baccianella-LREC-2010}. In text categorization by topic and authorship attribution, masking a token is conditioned by its part-of-speech tag. While content words receive higher masking probabilities for text categorization by topic, function words and punctuation tokens are more likely to be masked in authorship attribution \cite{Kestemont-CLFL-2014}.

We conduct fine-tuning experiments on four data sets: SST-2 \cite{Socher-EMNLP-2013}, 20 Newsgroups \cite{Lang-ICML-1995}, Reuters-21578 \cite{Lewis-UCI-1987}, and PAN 2019 Cross-Domain Authorship Attribution \cite{Kestemont-CLEF-2019}. Our experiments cover a diverse set of downstream tasks, including sentiment analysis, text classification by topic, and authorship attribution. We compare with several baselines, including conventional fine-tuning and standard MLM, and state-of-the-art training strategies based on anti-curriculum learning \cite{Ankner-EACL-2024} as well as curriculum learning \cite{Poesina-ACL-2024}. The results show that our strategy outperforms its competitors across all data sets. Moreover, TIACBM brings statistically significant performance gains, showing that harnessing task knowledge to mask tokens during fine-tuning is beneficial for multiple downstream tasks.

In summary, our contribution is threefold:
\begin{itemize}
    \item \vspace{-0.2cm} We propose to leverage task-specific knowledge to determine the probability distribution used to mask input tokens in MLM.
    \item \vspace{-0.2cm} We introduce a cyclic decaying masking ratio that boosts performance over existing anti-curriculum learning strategies for MLM.
    \item \vspace{-0.2cm} We apply a dynamic MLM strategy to fine-tune pre-trained models on downstream tasks, showing that MLM is not only beneficial for pre-training, but also for the fine-tuning stage.
\end{itemize}

\section{Related Work}
\label{sec:appendix1}
\vspace{-0.1cm}

Our framework is mostly related to work on curriculum learning \cite{Bengio-ICML-2009}. \citet{Soviany-IJCV-2022} divide curriculum learning methods into data-level \cite{Chang-EACL-2021,Gong-CIKM-2021,Kocmi-RANLP-2017,Liu-IJCAI-2018,Nagatsuka-NGC-2023}, model-level \cite{Croitoru-IJCV-2025,Sinha-NeurIPS-2020}, task-level \cite{Liu-IJCAI-2020,Narvekar-AAMAS-2016}, and objective-level \cite{Pathak-ICMLA-2019} strategies. Most of existing studies focus on computer vision \cite{Croitoru-IJCV-2025,Huang-CVPR-2020,Sinha-NeurIPS-2020} and reinforcement learning \cite{Fang-NeurIPS-2019,Florensa-CoRL-2017}. Methods in these domains are vaguely related to our approach, with some exceptions that employ curriculum based on masked image modeling (MIM) \cite{Jarca-ECAI-2024,Madan-WACV-2024}. In the image domain, the MIM approach was explored from multiple perspectives, which led to the development of adaptive masking strategies based on curriculum learning \cite{Madan-WACV-2024}, that can produce more robust representations. A notable finding is that an easy-to-hard curriculum works generally well for image masking \cite{Jarca-ECAI-2024}. In contrast, analogous studies focusing on text \cite{Ankner-EACL-2024,Yang-ACL-2023} suggest that a hard-to-easy curriculum, i.e.~using a decaying masking ratio, is more appropriate for text. Our results confirm the observations of \citet{Ankner-EACL-2024} and \citet{Yang-ACL-2023}, although we reset the masking ratio during training, resulting in a cyclic decaying masking ratio.

Curriculum learning methods specifically designed for text \cite{Gong-CIKM-2021,Kocmi-RANLP-2017,Liu-IJCAI-2018,Liu-ACL-2020,Zhan-AAAI-2021} are also related to our work. Some of the most popular approaches rely on text length \cite{Nagatsuka-NGC-2023} and model competence \cite{Platanios-NAACL-2019} to organize the samples from easy to hard. Recent approaches are based on more complex strategies. For instance, the state-of-the-art curriculum learning method proposed by \citet{Poesina-ACL-2024} employs data cartography \cite{Swayamdipta-EMNLP-2020} while training a baseline model to obtain the variability and confidence of each sample. The training data is further mapped as easy, ambiguous or hard. The model is then retrained via an easy-to-hard curriculum. To boost performance, the method employs stratified sampling as well as a continuous function to map the data points, resulting in a method called Cart-Stra-CL++.

Different from related curriculum and anti-curriculum learning methods \cite{Ankner-EACL-2024,Poesina-ACL-2024,Yang-ACL-2023}, we design an anti-curriculum strategy for the fine-tuning stage of pre-trained language models, leveraging knowledge about the downstream tasks. Moreover, our novel design leads to superior performance on a range of downstream tasks.

\section{Method}
\vspace{-0.1cm}

We propose a novel task-informed anti-curriculum by masking to fine-tune pre-trained language models. Specifically, we employ a cyclic decaying masking ratio which encourages a progressive adaptation of the model over time. In addition, we harness task-specific knowledge to determine which tokens need to be masked, ensuring that the model focuses on the most relevant words in a sentence. By combining a dynamic masking ratio with selective token masking, our strategy can boost performance on complex downstream tasks. We emphasize that relevant words are those that lead to discriminative features. However, deep neural networks can learn co-adapted features, which affects generalization capacity. Masking some of the discriminative features prevents feature co-adaptation \cite{Hinton-Arxiv-2012}. TIACBM achieves this effect in a targeted manner, by strategically starting with a higher masking ratio and gradually reducing it, which aligns with theories of curriculum learning and adversarial training.

Prior to the start of the training, we create a vector $\mathbf{r}=\{r_1 \geq \dots \geq r_K\} \in [0,1]^K$ containing $K$ masking ratios, which represents the anti-curriculum schedule. Note that $K \ll T$, where $T$ is the total number of training iterations. Thus, after every $K$ iterations, we reuse the masking ratios starting with $r_1$. In Algorithm~\ref{masking_alg}, we formally present how the masking is performed for a training text sample, given as a sequence of tokens $\mathbf{x}$. In the first step, we compute the number of tokens to be masked $N$, based on the sequence length and the masking ratio $r_t$ of the current training iteration $t$. In the second step, for each token $x_i$, we call a task-specific function to compute its importance, taking into account the token and the surrounding text. Additionally, in this step, we also normalize the importance scores to obtain probability values. Finally, in the third and last step, we build a categorical distribution from these probabilities and sample $N$ tokens to mask. We further describe and motivate the $\verb|task_relevance|$ function for each task.

\setlength{\textfloatsep}{8pt}
\SetNlSty{textt}{}{.}
\SetCommentSty{mycommfont}
\begin{algorithm}[!t]
\small{
\caption{{TIACBM}}
\label{masking_alg}
\SetAlgoLined
\SetKwInOut{Input}{Input}  
\textbf{Input}: $\mathbf{x}$ -- training sequence of tokens;\\
$\mathbf{r} = \{r_1, \dots, r_K\}$ -- decaying masking ratio schedule;\\
$\verb|task_relevance|$ -- function that computes the task-specific importance of a word; \\
$t$ -- current training iteration.\\
\textbf{Output}: $\hat{\mathbf{x}}$ -- masked sequence of tokens. \\
{\color{blue}{\tcc{1: Determine the number of masked tokens.}}}\nl
$r_t \gets \mathbf{r}[t \mod K]$;\\
$N \gets \lfloor \left|\mathbf{x}\right| \cdot r_t \rfloor$;\\
{\color{blue}{\tcc{2: Compute the task-specific importance and normalize the values to obtain a probability distribution over all tokens.}}}\nl
$\mathbf{s} \gets \{s_i = \verb|task_relevance|(x_i, \mathbf{x}), \forall i=1, \dots, \left|\mathbf{x}\right| \}$;\\
$\mathbf{p} \gets \left\{ p_i =  \frac{s_i}{\Sigma_{j=1}^{\left|\mathbf{s}\right|} s_j}, \forall i=1, \dots, \left|\mathbf{s}\right|\right\}$\\
{\color{blue}{\tcc{3: Mask the tokens.}}}\nl
$n \gets 0$;\\
$\hat{\mathbf{x}} \gets \mathbf{x}$; \\

\While{$n < N $}{
        $ i ~\sim \mathrm{Categorical}(\left|\mathbf{x}\right|, \mathbf{p})$;\\
        \If{$\hat{x}_i \notin \{\mbox{\texttt{[MASK], [SEP], [CLS]}}\}$}
        {
            $\hat{x}_i \gets \mbox{\texttt{[MASK]}}$; \\
            $n \gets n + 1$;
        }
}       
\nl \textbf{return} $\hat{\mathbf{x}}$ 
}
\end{algorithm}

\noindent
\textbf{Sentiment analysis.}
For polarity classification, we hypothesize that the most subjective words represent the most important features, and masking them will result in a hard-to-easy curriculum. The core foundation of this approach lies in using the SentiWordNet 3.0 sentiment lexicon~\citep{Baccianella-LREC-2010}. For each sentence, we analyze the most probable synset of each word, using a generic Lesk algorithm, and search for it in the lexicon. 
This process aims to determine the most likely positive $\left(s_{\text{pos}}\right)$ and negative scores $\left(s_{\text{neg}}\right)$ for the current word. Both scores range from 0 to 1, with higher values indicating stronger positive or negative connotations, respectively. We emphasize that these scores are linked together and their sum is lower or equal to $1$. Based on these values, \citet{Baccianella-LREC-2010} determine the objectivity score for each word as: 
\begin{equation}
    \label{objectivity_score}
    o = 1 - (s_{\text{pos}} + s_{\text{neg}}).
\end{equation}
In contrast, we leverage the subjectivity score as an importance measure to form the vector $\mathbf{s}$ for a given input $\mathbf{x}$ in Algorithm~\ref{masking_alg}. Accordingly, we compute the importance score $s_i$ for each token $x_i$ as follows:
\begin{equation}
    \label{sent_importance_score}
    s_i =  (s_{\text{pos}}^i + s_{\text{neg}}^i), \forall i = 1,\dots, \left|\mathbf{x}\right|.
\end{equation}

\noindent
\textbf{Text categorization.}
For text categorization, we hypothesize that content words (nouns, verbs, adjectives, adverbs, and proper names) are more relevant. Consequently, we assign an importance score of 0 to every other part of speech. To compute the importance scores for content words of a given sequence, we also draw upon the knowledge of the pre-trained language model. Before fine-tuning, we extract the attention weights from each attention block and each attention head, given by:
\begin{equation}
    \label{eq_attention_weights}
    A_b^h =  \mbox{softmax}\left(\frac{Q_b^h \cdot(K_b^h)^\top}{\sqrt{d}}\right),
\end{equation}
where $b$ iterates over the self-attention blocks, $h$ iterates over the attention heads, and $Q_b^h \in \mathbb{R}^{\left|\mathbf{x}\right| \times d}$ and $K_b^h \in \mathbb{R}^{\left|\mathbf{x}\right| \times d}$ are the query and key matrices of the attention block $b$ and head $h$. The result, $A_b^h \in \mathbb{R}^{\left|\mathbf{x}\right| \times \left|\mathbf{x}\right|}$, is a square matrix containing the similarity between each token found in the input sequence $\mathbf{x}$ and all the other tokens. We further compute an importance vector $\mathbf{a}$ by averaging the attention matrices, as follows:
\begin{equation}
    \label{cls_content_importance_score}
    \mathbf{a} = \frac{1}{B\cdot H \cdot \left|\mathbf{x}\right|} \sum_{h=1}^{H} \sum_{b=1}^B\sum_{j=1}^{\left|\mathbf{x}\right|}{A_{b,j}^h}, 
\end{equation}
where $B$ is the number of attention blocks and $H$ is the number of heads. The final importance scores $\mathbf{s}$ employed in Algorithm~\ref{masking_alg} are computed as:
\begin{equation}
    \label{authorship_importance_score}
    s_i\!=\!\begin{cases}
        a_i, \text{if } x_i \text{ is a content word}\!\!\!\!\\
        0,  \text{otherwise}
    \end{cases}
     \!\!, \forall i\!=\!1, ...,\!\left|\mathbf{x}\right|.
\end{equation}

\noindent
\textbf{Authorship attribution.}
For the authorship attribution task, we compute the importance score vector $\mathbf{s}$ using a procedure similar to the one described for text classification. However, instead of using the content words for masking, we mask functional words, such as adpositions, determiners, conjunctions, symbols, particles, and punctuation. Authors exhibit consistent writing style patterns, which are reflected in their use of these functional words \cite{Kestemont-CLFL-2014}. Consequently, Eq.~\eqref{authorship_importance_score} is modified as follows:
\begin{equation}
    \label{cls_importance_score}
    s_i\!=\!\begin{cases}
        a_i, \text{if } x_i \text{ is a functional word}\!\!\!\!\\
        0,  \text{otherwise}
    \end{cases}
    \!\!, \forall i\!=\!1, ...,\!\left|\mathbf{x}\right|,
\end{equation}
where $a_i$ is computed as in Eq.~\eqref{cls_content_importance_score}.

\section{Experiments and Results}
\vspace{-0.1cm}
\subsection{Data sets}
\vspace{-0.1cm}

We evaluate TIACBM on the three tasks, namely sentiment analysis, text categorization and authorship 
attribution. 

\noindent
\textbf{Reuters-21578.} Reuters-21578~\cite{Lewis-UCI-1987} is a multi-label text categorization data set containing 12,449 training and 5,458 test instances. The data set gathers documents from 90 categories.

\noindent
\textbf{20 Newsgroups.} 20 Newsgroups~\cite{Lang-ICML-1995} is a multi-class data set for text categorization, which comprises 11,314 training instances and 7,532 test instances belonging to 20 classes. 

\noindent
\textbf{SST2.} The SST2 data set \cite{Socher-EMNLP-2013} is a popular benchmark for sentiment analysis, comprising 67,349 training and 872 validation samples, which are labeled either as positive or negative.

\noindent
\textbf{PAN19.} For authorship attribution, we use the PAN19~\cite{Kestemont-CLEF-2019} data set. We report results for Problem 0001 (P1) and Problem 0005 (P5). We discard the unknown files when computing the evaluation metrics. Both problems have 9 authors, with 7 training files each. There are 561 test files for P1, and 264 test files for P5. 

\begin{table*}[t]
  \centering
  \setlength\tabcolsep{0.12em}
  \small{
  \begin{tabular}{|l|l|c|c|c|c|c|c|c|}

     \hline
      \multirow{2}{*}{Model} & \multicolumn{1}{c|}{Fine-Tuning}  & Reuters & 20 News & SST2 & \multicolumn{2}{|c|}{PAN19-P1} & \multicolumn{2}{c|}{PAN19-P5}\\
     \cline{3-9}
     & \multicolumn{1}{c|}{Strategy} & \multicolumn{2}{c|}{Micro F1} & Accuracy & Accuracy & Macro F1 & Accuracy & Macro F1\\
    \hline
    \hline
    \multirow{6}{*}{BERT$_{\text{base}}$} 
       &
       Conventional & $90.61_{\pm0.28}$ &$84.63_{\pm0.28}$   & $93.38_{\pm0.14}$ & $69.76_{\pm15.11}$ & $58.24_{\pm6.06}$&$66.10_{\pm1.56}$ & $36.10_{\pm4.22}$\\

    &  Constant & $90.81_{\pm0.24}$& $84.98_{\pm0.12}$ & $93.94_{\pm0.15}$ & $63.70_{\pm9.96}$ & $47.50_{\pm7.26}$&$65.76_{\pm2.92}$ & $36.30_{\pm4.56}$\\
             
    &  \citet{Poesina-ACL-2024} & $90.72_{\pm0.13}$ & $82.30_{\pm0.25}$ & $94.00_{\pm0.14}$ &$55.23_{\pm3.48}$ & $44.76_{\pm2.56}$&$68.66_{\pm2.00}$ & $40.72_{\pm2.64}$\\
    
     &  \citet{Ankner-EACL-2024} & $90.99_{\pm0.05}$& $85.39_{\pm0.23}$ & $93.83_{\pm0.20}$ & $54.38_{\pm16.10}$ & $46.03_{\pm8.53}$&$65.55_{\pm0.85}$ & $36.60_{\pm4.28}$\\

    &  Cyclic Decaying & $90.96_{\pm0.12}$& $84.88_{\pm0.08}$ & $94.10_{\pm0.20}$&$50.56_{\pm15.22}$ & $51.94_{\pm8.99}$&$69.28_{\pm3.27}$ & $42.94_{\pm2.45}$\\

     &  TIACBM (ours) & $\mathbf{91.20}_{\pm0.20}$&  $\mathbf{85.65}_{\pm0.10}$& $\mathbf{94.61}_{\pm0.08}$ &$\mathbf{77.32}_{\pm9.33}$ & $\mathbf{60.60}_{\pm7.37}$ & $\mathbf{69.94}_{\pm1.98}$ & $\mathbf{44.20}_{\pm2.67}$\\
    \hline
       \multirow{6}{*}{RoBERTa$_{\text{base}}$} &
       Conventional & $90.55_{\pm0.18}$ & $84.49_{\pm0.11}$ &$94.56_{\pm0.09}$ &$89.20_{\pm3.01}$ & $76.92_{\pm4.64}$&$67.42_{\pm2.90}$ & $38.30_{\pm4.32}$\\

     &  Constant &$90.49_{\pm0.11}$ &$85.10_{\pm0.30}$  &$94.88_{\pm0.26}$ &$92.44_{\pm0.51}$ & $78.84_{\pm2.61}$&$64.00_{\pm4.33}$ & $33.62_{\pm5.41}$\\

    &  \citet{Poesina-ACL-2024}  &$90.52_{\pm0.14}$  &$79.89_{\pm0.34}$ &$94.81_{\pm0.23}$&$90.18_{\pm1.03}$ & $78.70_{\pm2.13}$&$65.16_{\pm1.26}$ & $33.36_{\pm0.95}$\\
         
     &  \citet{Ankner-EACL-2024} &$90.42_{\pm0.09}$ &$85.33_{\pm0.17}$  &$94.24_{\pm0.19}$ &$91.76_{\pm1.53}$ & $80.50_{\pm2.07}$&$65.10_{\pm0.49}$ & $37.64_{\pm2.78}$\\

     &  Cyclic Decaying &$90.70_{\pm0.14}$ &$84.74_{\pm0.20}$ &$94.70_{\pm0.14}$ &$91.36_{\pm1.24}$ & $78.84_{\pm2.21}$&$67.80_{\pm2.89}$ & $39.36_{\pm2.63}$\\
    
     &  TIACBM (ours) &$\mathbf{91.06}_{\pm0.19}$ &$\mathbf{85.93}_{\pm0.18}$  &$\mathbf{95.04}_{\pm0.18}$ &$\mathbf{93.98}_{\pm0.70}$ & $\mathbf{83.78}_{\pm2.55}$ & $\mathbf{68.38}_{\pm1.53}$ & $\mathbf{41.86}_{\pm2.15}$\\
    \hline
  \end{tabular}
  }\vspace{-0.25cm}
    \caption{Results on text classification (Reuters-21578, 20 Newsgroups), sentiment analysis (SST2) and authorship attribution (PAN19), with BERT and RoBERTa. Cochran's Q testing confirms that the results of TIACBM are always statistically better than conventional fine-tuning (p-value$<0.001$). The top scores for each architecture and metric are highlighted in bold.}
  \label{tab_strategies}
  \vspace{-0.15cm}
\end{table*}

\begin{table*}[t]
  \centering
  \setlength\tabcolsep{0.3em}
  \small{
  \begin{tabular}{|l|l|c|c|c|c|c|}
    \hline
    \multirow{2}{*}{Model} &  \multirow{2}{*}{Fine-Tuning Strategy}  & \multicolumn{1}{c|}{SST2} & \multicolumn{2}{c|}{PAN19-P1} & \multicolumn{2}{c|}{PAN19-P5} \\
    \cline{3-7}
    & & Accuracy & Accuracy & Macro F1 & Accuracy & Macro F1 \\
    \hline
    \hline
    \multirow{6}{*}{GPT-2}
      & Conventional       & $92.35_{\pm0.30}$ & $82.78_{\pm3.06}$ & $67.96_{\pm3.98}$ & $62.88_{\pm3.64}$ & $38.16_{\pm3.93}$ \\
      & Constant           & $92.54_{\pm0.43}$ & $59.10_{\pm1.81}$ & $41.58_{\pm0.76}$ & $54.54_{\pm6.95}$ & $32.08_{\pm2.30}$ \\
      & \citet{Poesina-ACL-2024} & $92.27_{\pm0.21}$ & $72.44_{\pm6.97}$ & $57.60_{\pm7.02}$ & $59.54_{\pm6.95}$ & $32.86_{\pm2.94}$ \\
      & \citet{Ankner-EACL-2024}  & $92.60_{\pm0.09}$ & $83.76_{\pm1.46}$ & $68.70_{\pm2.13}$ & $62.42_{\pm0.39}$ & $36.90_{\pm2.40}$ \\
      & Cyclic Decaying   & $92.74_{\pm0.05}$ & $82.38_{\pm3.51}$ & $67.34_{\pm4.57}$ & $65.00_{\pm3.92}$ & $36.46_{\pm2.81}$ \\
      
      & TIACBM (ours) & $\mathbf{92.96}_{\pm0.14}$ & $\mathbf{85.90}_{\pm1.96}$ & $\mathbf{73.44}_{\pm3.81}$ & $\mathbf{68.20}_{\pm2.25}$ & $\mathbf{42.90}_{\pm2.56}$ \\
    \hline
  \end{tabular}
  }\vspace{-0.25cm}
  \caption{\label{tab:sst2_pan19_gpt2}Results on sentiment analysis (SST2) and authorship attribution (PAN19), with GPT-2. Cochran's Q testing confirms that the results of TIACBM are always statistically better than conventional fine-tuning (p-value$<0.001$). The top score for each metric is highlighted in bold.}
  \vspace{-0.25cm}
\end{table*}

\subsection{Baselines}
\label{experiments}
\vspace{-0.1cm}

We compare TIACBM with five fine-tuning strategies, which are described in detail below.

\noindent
\textbf{Conventional.} This is the standard fine-tuning approach, which does not involve MLM. It uses the CB-NTR loss \cite{Huang-EMNLP-2021} for Reuters-21578, due to its long-tail distribution.

\noindent
\textbf{Constant.} This fine-tuning strategy uses a constant masking ratio to mask input tokens. The masking ratio is set to 15\%, following \citet{Devlin-NAACL-2019}.

\noindent
\textbf{Cart-Stra-CL++.} This is a state-of-the-art easy-to-hard curriculum approach introduced by \citet{Poesina-ACL-2024}. This method needs to perform data cartography for the baseline fine-tuned with the conventional regime, before employing the curriculum. This essentially doubles the training time.

\noindent
\textbf{Decaying Masking Ratio.} The decaying masking ratio, a.k.a.~anti-curriculum by masking, is proposed by \citet{Ankner-EACL-2024}. This training strategy can be seen as an ablated version of our approach, which is obtained by dropping the cyclical regime ($K=T$) and by discarding task-specific information.

\noindent
\textbf{Cyclic Decaying Masking Ratio.} This is an ablated version of our approach, which simply discards the task-specific information.

\subsection{Experimental Setup}
\vspace{-0.1cm}

We employ two pre-trained masked language models, BERT$_{\text{base}}$ \cite{Devlin-NAACL-2019} and RoBERTa$_{\text{base}}$ \cite{Liu-ICLR-2020}, in order to evaluate the various fine-tuning strategies. We also experiment with GPT-2 \cite{Radford-OAI-2019} to show that TIACBM can be applied beyond masked language models.

We run all experiments three times and report the mean and standard deviation for each experiment. We execute the fine-tuning for $15$ epochs for sentiment analysis, using a learning rate of $5\cdot 10^{-5}$, a batch size of $64$ and a max token length of $100$. For text categorization, we train for $30$ epochs, using a learning rate of $10^{-4}$, a batch size of $32$ and a max token length of $512$. For authorship attribution, we use $30$ epochs, a batch size of $8$ and a max token length of $512$. For PAN19-P1, we use a learning rate of $10^{-4}$ for BERT and $10^{-5}$ for RoBERTa and GPT-2. For PAN19-P5, we set the learning rate to $5\cdot 10^{-5}$ for all language models. We use the cross-entropy loss for all data sets, except on Reuters-21578. In this case, the baseline language model is fine-tuned with the CB-NTR loss \cite{Huang-EMNLP-2021}. We keep the same loss for all fine-tuning strategies on Reuters-21578. In terms of optimizers, we use Adamax for the sentiment analysis task, and AdamW for the others. We keep these parameters consistent across all baseline methods and TIACBM. We release our code to reproduce the results at \url{https://github.com/JarcaAndrei/TIACBM}.

\subsection{Results}
\vspace{-0.1cm}

We compare our approach against the competing fine-tuning strategies in Table~\ref{tab_strategies}. Our method improves performance across all downstream tasks, while exhibiting lower variability. Moreover, TIACBM brings significant gains across both BERT and RoBERTa. On SST2, TIACBM provides an increase of $1.23\%$ over the baseline, and $0.61\%$ over the state-of-the-art method of \citet{Poesina-ACL-2024}, suggesting that masking the harder (subjective) words towards the easier (objective) words, in a cyclic fashion, improves the performance of the model on sentiment analysis. On Reuters-21578, our approach boosts the micro $F_1$ score by $0.59\%$, reaching a top result of $91.48\%$, surpassing the state-of-the-art model based on CB-NTR \cite{Huang-EMNLP-2021}. On 20 Newsgroups, TIACBM outperforms the baseline by $1.44\%$ and Cart-Stra-CL++ \cite{Poesina-ACL-2024} by $6.04\%$, highlighting the effectiveness of content word masking. For PAN19, TIACBM increases both accuracy and macro $F_1$ score by $2.36\%$ and $15.84\%$, respectively, demonstrating robust generalization even with limited training data.
Compared with the approach of \citet{Ankner-EACL-2024}, we observe that TIACM brings higher performance gains across all tasks, mainly due to the task-specific information harnessed by our approach.

We present additional experiments with GPT-2 in Table \ref{tab:sst2_pan19_gpt2}. The results obtained with GPT-2 are consistent with those obtained with BERT and RoBERTa, confirming that TIACBM leads to significant performance gains. Hence, the results reported in Table \ref{tab:sst2_pan19_gpt2} indicate that TIACBM is not limited to masked language models, being a generic approach that can be applied to any LLM.

\section{Conclusion}
\vspace{-0.1cm}
We proposed a novel task-informed anti-curriculum by masking approach (TIACBM), and we evaluated its effectiveness on three tasks: sentiment analysis, text categorization by topic, and authorship attribution. The proposed method leverages information about the downstream tasks to decide which tokens to select for masking in a novel anti-curriculum by masking framework.
On all the three tasks, our method achieved better results across all experiments, outperforming both baselines and state-of-the-art methods. Moreover, our method performed well on both multi-label and multi-class classification, while also proving resilience against imbalanced data sets, such as Reuters-21578. Additionally, we also showed that TIACBM is effective in scenarios with a low number of training samples, as in the case of PAN19.

\noindent
\textbf{Acknowledgments.}
This research is supported by the project ``Romanian Hub for Artificial Intelligence - HRIA'', Smart Growth, Digitization and Financial Instruments Program, 2021-2027, MySMIS no. 334906.

\section{Limitations} 

We present a novel method to consistently improve the performance of language models on downstream tasks. However, there is no universal anti-curriculum (masking ratio) schedule that can work for all models or data sets, representing an important parameter to be optimized by the user. Still, a general empirical statement proven in our work is that cycling anti-curriculum schedulers are superior in NLP, when it comes to curriculum by masking. Additionally, our method computes token relevance using a task-specific approach and this can be challenging to design for some tasks. We show on two tasks that attention weights can effectively serve this purpose.

\section{Ethics Statement}
To our knowledge, the proposed method poses no immediate risk. However, it can be adapted for generative modeling, which raises concerns about its potential misuse for malicious purposes, such as fake content generation.

\bibliography{custom}

\begin{thebibliography}{37}
\providecommand{\natexlab}[1]{#1}

\bibitem[{Ankner et~al.(2024)Ankner, Saphra, Blalock, Frankle, and
  Leavitt}]{Ankner-EACL-2024}
Zachary Ankner, Naomi Saphra, Davis Blalock, Jonathan Frankle, and Matthew
  Leavitt. 2024.
\newblock \href {https://aclanthology.org/2024.eacl-short.42/} {Dynamic masking
  rate schedules for {MLM} pretraining}.
\newblock In \emph{Proceedings of the 18th Conference of the European Chapter
  of the Association for Computational Linguistics (EACL)}, pages 477--487.
  Association for Computational Linguistics.

\bibitem[{Baccianella et~al.(2010)Baccianella, Esuli, and
  Sebastiani}]{Baccianella-LREC-2010}
Stefano Baccianella, Andrea Esuli, and Fabrizio Sebastiani. 2010.
\newblock \href {https://aclanthology.org/L10-1531/} {{{S}enti{W}ord{N}et 3.0:
  An Enhanced Lexical Resource for Sentiment Analysis and Opinion Mining}}.
\newblock In \emph{Proceedings of the Seventh International Conference on
  Language Resources and Evaluation (LREC)}, pages 2200--2024. European
  Language Resources Association.

\bibitem[{Bengio et~al.(2009)Bengio, Louradour, Collobert, and
  Weston}]{Bengio-ICML-2009}
Yoshua Bengio, J{\'e}r\^{o}me Louradour, Ronan Collobert, and Jason Weston.
  2009.
\newblock \href {https://doi.org/10.1145/1553374.1553380} {Curriculum
  learning}.
\newblock In \emph{Proceedings of the International Conference on Machine
  Learning (ICML)}, pages 41--48. ACM.

\bibitem[{Chang et~al.(2021)Chang, Yeh, and Demberg}]{Chang-EACL-2021}
Ernie Chang, Hui-Syuan Yeh, and Vera Demberg. 2021.
\newblock \href {https://doi.org/10.18653/v1/2021.eacl-main.61} {Does the order
  of training samples matter? improving neural data-to-text generation with
  curriculum learning}.
\newblock In \emph{Proceedings of the 16th Conference of the European Chapter
  of the Association for Computational Linguistics (EACL)}, pages 727--733.
  Association for Computational Linguistics.

\bibitem[{Croitoru et~al.(2025)Croitoru, Ristea, Ionescu, and
  Sebe}]{Croitoru-IJCV-2025}
Florinel-Alin Croitoru, Nicolae-C{\u{a}}t{\u{a}}lin Ristea, Radu~Tudor Ionescu,
  and Nicu Sebe. 2025.
\newblock \href {https://doi.org/10.1007/s11263-024-02186-5} {Learning rate
  curriculum}.
\newblock \emph{International Journal of Computer Vision}, 133(1):291--314.

\bibitem[{Devlin et~al.(2019)Devlin, Chang, Lee, and
  Toutanova}]{Devlin-NAACL-2019}
Jacob Devlin, Ming-Wei Chang, Kenton Lee, and Kristina Toutanova. 2019.
\newblock \href {https://www.aclweb.org/anthology/N19-1423} {{BERT:
  Pre-training of Deep Bidirectional Transformers for Language Understanding}}.
\newblock In \emph{Proceedings of the 2019 Conference of the North American
  Chapter of the Association for Computational Linguistics: Human Language
  Technologies (NAACL-HLT)}, pages 4171--4186. Association for Computational
  Linguistics.

\bibitem[{Fang et~al.(2019)Fang, Zhou, Du, Han, and Zhang}]{Fang-NeurIPS-2019}
Meng Fang, Tianyi Zhou, Yali Du, Lei Han, and Zhengyou Zhang. 2019.
\newblock \href
  {https://proceedings.neurips.cc/paper_files/paper/2019/file/83715fd4755b33f9c3958e1a9ee221e1-Paper.pdf}
  {{Curriculum-guided Hindsight Experience Replay}}.
\newblock In \emph{Proceedings the 33rd International Conference on Neural
  Information Processing Systems (NeurIPS)}, pages 12623--12634. Curran
  Associates, Inc.

\bibitem[{Florensa et~al.(2017)Florensa, Held, Wulfmeier, Zhang, and
  Abbeel}]{Florensa-CoRL-2017}
Carlos Florensa, David Held, Markus Wulfmeier, Michael Zhang, and Pieter
  Abbeel. 2017.
\newblock \href {http://proceedings.mlr.press/v78/florensa17a/florensa17a.pdf}
  {Reverse curriculum generation for reinforcement learning}.
\newblock In \emph{Proceedings of the 1st Annual Conference on Robot Learning
  (CoRL)}, volume~78, pages 482--495. PMLR.

\bibitem[{Gong et~al.(2021)Gong, Liu, Yuan, Yang, Cai, Wan, Chen, Niu, and
  Wang}]{Gong-CIKM-2021}
Yantao Gong, Cao Liu, Jiazhen Yuan, Fan Yang, Xunliang Cai, Guanglu Wan,
  Jiansong Chen, Ruiyao Niu, and Houfeng Wang. 2021.
\newblock \href {https://doi.org/10.1145/3459637.3482082} {Density-based
  dynamic curriculum learning for intent detection}.
\newblock In \emph{Proceedings of the 30th ACM International Conference on
  Information \& Knowledge Management (CIKM)}, pages 3034--3037.

\bibitem[{Hinton et~al.(2012)Hinton, Srivastava, Krizhevsky, Sutskever, and
  Salakhutdinov}]{Hinton-Arxiv-2012}
Geoffrey~E. Hinton, Nitish Srivastava, Alex Krizhevsky, Ilya Sutskever, and
  Ruslan~R Salakhutdinov. 2012.
\newblock \href {https://arxiv.org/abs/1207.0580} {Improving neural networks by
  preventing co-adaptation of feature detectors}.
\newblock \emph{arXiv preprint arXiv:1207.0580}.

\bibitem[{Huang et~al.(2021)Huang, Giledereli, K{\"o}ksal, {\"O}zg{\"u}r, and
  Ozkirimli}]{Huang-EMNLP-2021}
Yi~Huang, Buse Giledereli, Abdullatif K{\"o}ksal, Arzucan {\"O}zg{\"u}r, and
  Elif Ozkirimli. 2021.
\newblock \href {https://doi.org/10.18653/v1/2021.emnlp-main.643} {Balancing
  methods for multi-label text classification with long-tailed class
  distribution}.
\newblock In \emph{Proceedings of the 2021 Conference on Empirical Methods in
  Natural Language Processing (EMNLP)}, pages 8153--8161. Association for
  Computational Linguistics.

\bibitem[{Huang et~al.(2020)Huang, Wang, Tai, Liu, Shen, Li, Li, and
  Huang}]{Huang-CVPR-2020}
Yuge Huang, Yuhan Wang, Ying Tai, Xiaoming Liu, Pengcheng Shen, Shaoxin Li,
  Jilin Li, and Feiyue Huang. 2020.
\newblock \href {https://doi.org/10.1109/CVPR42600.2020.00594}
  {{CurricularFace: Adaptive Curriculum Learning Loss for Deep Face
  Recognition}}.
\newblock In \emph{Proceedings of IEEE/CVF Conference on Computer Vision and
  Pattern Recognition (CVPR)}, pages 5900--5909. IEEE.

\bibitem[{Jarca et~al.(2024)Jarca, Croitoru, and Ionescu}]{Jarca-ECAI-2024}
Andrei Jarca, Florinel{-}Alin Croitoru, and Radu~Tudor Ionescu. 2024.
\newblock \href {https://doi.org/10.3233/FAIA240503} {{{CBM:} Curriculum by
  Masking}}.
\newblock In \emph{Proceedings of the 27th European Conference on Artificial
  Intelligence (ECAI)}, volume 392, pages 314--321. {IOS} Press.

\bibitem[{Kestemont(2014)}]{Kestemont-CLFL-2014}
Mike Kestemont. 2014.
\newblock \href {https://doi.org/10.3115/v1/W14-0908} {{Function Words in
  Authorship Attribution. From Black Magic to Theory?}}
\newblock In \emph{Proceedings of the 3rd Workshop on Computational Linguistics
  for Literature ({CLFL})}, pages 59--66. Association for Computational
  Linguistics.

\bibitem[{Kestemont et~al.(2019)Kestemont, Stamatatos, Manjavacas, Daelemans,
  Potthast, and Stein}]{Kestemont-CLEF-2019}
Mike Kestemont, Efstathios Stamatatos, Enrique Manjavacas, Walter Daelemans,
  Martin Potthast, and Benno Stein. 2019.
\newblock \href {https://ceur-ws.org/Vol-2380/paper_264.pdf} {{Overview of the
  Cross-domain Authorship Attribution Task at PAN 2019}}.
\newblock In \emph{Working Notes Papers of the CLEF 2019 Evaluation Labs},
  volume 2380 of \emph{CEUR Workshop Proceedings}.

\bibitem[{Kocmi and Bojar(2017)}]{Kocmi-RANLP-2017}
Tom Kocmi and Ond{\v{r}}ej Bojar. 2017.
\newblock \href {https://doi.org/10.26615/978-954-452-049-6_050} {{Curriculum
  Learning and Minibatch Bucketing in Neural Machine Translation}}.
\newblock In \emph{Proceedings of the International Conference Recent Advances
  in Natural Language Processing (RANLP)}, pages 379--386. INCOMA Ltd.

\bibitem[{Lang(1995)}]{Lang-ICML-1995}
Ken Lang. 1995.
\newblock \href {https://doi.org/10.1016/B978-1-55860-377-6.50048-7}
  {{NewsWeeder: Learning to Filter Netnews}}.
\newblock In \emph{Proceedings of the Twelfth International Conference on
  Machine Learning (ICML)}, pages 331--339. Morgan Kaufmann Publishers Inc.

\bibitem[{Lewis(1987)}]{Lewis-UCI-1987}
David Lewis. 1987.
\newblock \href {https://doi.org/10.24432/C52G6M} {{Reuters-21578 Text
  Categorization Collection}}.
\newblock UCI Machine Learning Repository.

\bibitem[{Liu et~al.(2018)Liu, He, Liu, and Zhao}]{Liu-IJCAI-2018}
Cao Liu, Shizhu He, Kang Liu, and Jun Zhao. 2018.
\newblock \href {https://doi.org/10.24963/ijcai.2018/587} {{Curriculum Learning
  for Natural Answer Generation}}.
\newblock In \emph{Proceedings of the Twenty-Seventh International Joint
  Conference on Artificial Intelligence (IJCAI)}, pages 4223--4229.
  International Joint Conferences on Artificial Intelligence Organization.

\bibitem[{Liu et~al.(2022)Liu, Tian, Chen, Liu, Belagiannis, and
  Carneiro}]{Liu-CVPR-2022}
Fengbei Liu, Yu~Tian, Yuanhong Chen, Yuyuan Liu, Vasileios Belagiannis, and
  Gustavo Carneiro. 2022.
\newblock \href {https://doi.org/10.1109/CVPR52688.2022.02004} {{ACPL:
  Anti-curriculum Pseudo-labelling for Semi-supervised Medical Image
  Classification}}.
\newblock In \emph{Proceedings of the IEEE/CVF Conference on Computer Vision
  and Pattern Recognition (CVPR)}, pages 20697--20706. IEEE.

\bibitem[{Liu et~al.(2020{\natexlab{a}})Liu, Ren, Tan, Zhang, Qin, Zhao, and
  Liu}]{Liu-IJCAI-2020}
Jinglin Liu, Yi~Ren, Xu~Tan, Chen Zhang, Tao Qin, Zhou Zhao, and Tie{-}Yan Liu.
  2020{\natexlab{a}}.
\newblock \href {https://www.ijcai.org/proceedings/2020/0534.pdf} {Task-level
  curriculum learning for non-autoregressive neural machine translation}.
\newblock In \emph{Proceedings of the Twenty-Ninth International Conference on
  International Joint Conferences on Artificial Intelligence (IJCAI)}, pages
  3861--3867. International Joint Conferences on Artificial Intelligence
  Organization.

\bibitem[{Liu et~al.(2020{\natexlab{b}})Liu, Lai, Wong, and
  Chao}]{Liu-ACL-2020}
Xuebo Liu, Houtim Lai, Derek~F. Wong, and Lidia~S. Chao. 2020{\natexlab{b}}.
\newblock \href {https://doi.org/10.18653/v1/2020.acl-main.41} {Norm-based
  curriculum learning for neural machine translation}.
\newblock In \emph{Proceedings of the 58th Annual Meeting of the Association
  for Computational Linguistics (ACL)}, pages 427--436. Association for
  Computational Linguistics.

\bibitem[{Liu et~al.(2019)Liu, Ott, Goyal, Du, Joshi, Chen, Levy, Lewis,
  Zettlemoyer, and Stoyanov}]{Liu-ICLR-2020}
Yinhan Liu, Myle Ott, Naman Goyal, Jingfei Du, Mandar Joshi, Danqi Chen, Omer
  Levy, Mike Lewis, Luke Zettlemoyer, and Veselin Stoyanov. 2019.
\newblock \href {https://arxiv.org/abs/1907.11692} {{RoBERTa: A Robustly
  Optimized BERT Pretraining Approach}}.
\newblock In \emph{Proceedings of the International Conference on Learning
  Representations (ICLR)}.

\bibitem[{Madan et~al.(2024)Madan, Ristea, Nasrollahi, Moeslund, and
  Ionescu}]{Madan-WACV-2024}
Neelu Madan, Nicolae-C{\u{a}}t{\u{a}}lin Ristea, Kamal Nasrollahi, Thomas~B
  Moeslund, and Radu~Tudor Ionescu. 2024.
\newblock \href {https://doi.org/10.1109/WACV57701.2024.00248} {{CL-MAE:
  Curriculum-Learned Masked Autoencoders}}.
\newblock In \emph{Proceedings of the IEEE/CVF Winter Conference on
  Applications of Computer Vision (WACV)}, pages 2492--2502. IEEE.

\bibitem[{Nagatsuka et~al.(2023)Nagatsuka, Broni-Bediako, and
  Atsumi}]{Nagatsuka-NGC-2023}
Koichi Nagatsuka, Clifford Broni-Bediako, and Masayasu Atsumi. 2023.
\newblock \href {https://doi.org/10.1007/s00354-022-00198-8} {{Length-Based
  Curriculum Learning for Efficient Pre-training of Language Models}}.
\newblock \emph{New Generation Computing}, 41(1):109--134.

\bibitem[{Narvekar et~al.(2016)Narvekar, Sinapov, Leonetti, and
  Stone}]{Narvekar-AAMAS-2016}
Sanmit Narvekar, Jivko Sinapov, Matteo Leonetti, and Peter Stone. 2016.
\newblock \href {https://dl.acm.org/doi/pdf/10.5555/2936924.2937007} {Source
  task creation for curriculum learning}.
\newblock In \emph{Proceedings of the 2016 International Conference on
  Autonomous Agents \& Multiagent Systems (AAMAS)}, pages 566--574.
  International Foundation for Autonomous Agents and Multiagent Systems.

\bibitem[{Pathak and Paffenroth(2019)}]{Pathak-ICMLA-2019}
Harsh~Nilesh Pathak and Randy Paffenroth. 2019.
\newblock \href {https://doi.org/10.1109/ICMLA.2019.00268} {Parameter
  continuation methods for the optimization of deep neural networks}.
\newblock In \emph{Proceedings of the 18th IEEE International Conference On
  Machine Learning And Applications (ICMLA)}, pages 1637--1643.

\bibitem[{Platanios et~al.(2019)Platanios, Stretcu, Neubig, Poczos, and
  Mitchell}]{Platanios-NAACL-2019}
Emmanouil~Antonios Platanios, Otilia Stretcu, Graham Neubig, Barnabas Poczos,
  and Tom Mitchell. 2019.
\newblock \href {https://doi.org/10.18653/v1/N19-1119} {Competence-based
  curriculum learning for neural machine translation}.
\newblock In \emph{Proceedings of Proceedings of the 2019 Conference of the
  North {A}merican Chapter of the Association for Computational Linguistics:
  Human Language Technologies (NAACL-HLT)}, pages 1162--1172. Association for
  Computational Linguistics.

\bibitem[{Poesina et~al.(2024)Poesina, Caragea, and Ionescu}]{Poesina-ACL-2024}
Eduard~Gabriel Poesina, Cornelia Caragea, and Radu~Tudor Ionescu. 2024.
\newblock \href {https://doi.org/10.18653/v1/2024.acl-long.15} {{A Novel
  Cartography-Based Curriculum Learning Method Applied on RoNLI: The First
  Romanian Natural Language Inference Corpus}}.
\newblock In \emph{Proceedings of the 62nd Annual Meeting of the Association
  for Computational Linguistics (ACL)}, pages 236--253. Association for
  Computational Linguistics.

\bibitem[{Radford et~al.(2019)Radford, Wu, Child, Luan, Amodei, Sutskever
  et~al.}]{Radford-OAI-2019}
Alec Radford, Jeffrey Wu, Rewon Child, David Luan, Dario Amodei, Ilya
  Sutskever, et~al. 2019.
\newblock \href
  {https://cdn.openai.com/better-language-models/language_models_are_unsupervised_multitask_learners.pdf}
  {Language models are unsupervised multitask learners}.
\newblock \emph{OpenAI Blog}.

\bibitem[{Sinha et~al.(2020)Sinha, Garg, and Larochelle}]{Sinha-NeurIPS-2020}
Samarth Sinha, Animesh Garg, and Hugo Larochelle. 2020.
\newblock \href
  {https://proceedings.neurips.cc/paper/2020/file/f6a673f09493afcd8b129a0bcf1cd5bc-Paper.pdf}
  {Curriculum by smoothing}.
\newblock In \emph{Proceedings of the 34th International Conference on Neural
  Information Processing Systems (NeurIPS)}, pages 21653--21664. Curran
  Associates Inc.

\bibitem[{Socher et~al.(2013)Socher, Perelygin, Wu, Chuang, Manning, Ng, and
  Potts}]{Socher-EMNLP-2013}
Richard Socher, Alex Perelygin, Jean Wu, Jason Chuang, Christopher~D. Manning,
  Andrew Ng, and Christopher Potts. 2013.
\newblock \href {https://www.aclweb.org/anthology/D13-1170} {Recursive deep
  models for semantic compositionality over a sentiment treebank}.
\newblock In \emph{Proceedings of the 2013 Conference on Empirical Methods in
  Natural Language Processing (EMNLP)}, pages 1631--1642. Association for
  Computational Linguistics.

\bibitem[{Soviany et~al.(2022)Soviany, Ionescu, Rota, and
  Sebe}]{Soviany-IJCV-2022}
Petru Soviany, Radu~Tudor Ionescu, Paolo Rota, and Nicu Sebe. 2022.
\newblock \href {https://doi.org/10.1007/s11263-022-01611-x} {Curriculum
  learning: A survey}.
\newblock \emph{International Journal of Computer Vision}, 130:1526--1565.

\bibitem[{Swayamdipta et~al.(2020)Swayamdipta, Schwartz, Lourie, Wang,
  Hajishirzi, Smith, and Choi}]{Swayamdipta-EMNLP-2020}
Swabha Swayamdipta, Roy Schwartz, Nicholas Lourie, Yizhong Wang, Hannaneh
  Hajishirzi, Noah~A. Smith, and Yejin Choi. 2020.
\newblock \href {https://doi.org/10.18653/v1/2020.emnlp-main.746} {{Dataset
  Cartography: Mapping and Diagnosing Datasets with Training Dynamics}}.
\newblock In \emph{Proceedings of the 2020 Conference on Empirical Methods in
  Natural Language Processing (EMNLP)}, pages 9275--9293. Association for
  Computational Linguistics.

\bibitem[{Wettig et~al.(2023)Wettig, Gao, Zhong, and Chen}]{Wettig-ECAL-2023}
Alexander Wettig, Tianyu Gao, Zexuan Zhong, and Danqi Chen. 2023.
\newblock \href {https://doi.org/10.18653/v1/2023.eacl-main.217} {Should you
  mask 15{\%} in masked language modeling?}
\newblock In \emph{Proceedings of the 17th Conference of the European Chapter
  of the Association for Computational Linguistics (EACL)}, pages 2985--3000.
  Association for Computational Linguistics.

\bibitem[{Yang et~al.(2023)Yang, Zhang, and Zhao}]{Yang-ACL-2023}
Dongjie Yang, Zhuosheng Zhang, and Hai Zhao. 2023.
\newblock \href {https://doi.org/10.18653/v1/2023.acl-long.400} {Learning
  better masking for better language model pre-training}.
\newblock In \emph{Proceedings of the 61st Annual Meeting of the Association
  for Computational Linguistics (ACL)}, pages 7255--7267. Association for
  Computational Linguistics.

\bibitem[{Zhan et~al.(2021)Zhan, Liu, Wong, and Chao}]{Zhan-AAAI-2021}
Runzhe Zhan, Xuebo Liu, Derek~F Wong, and Lidia~S Chao. 2021.
\newblock \href {https://doi.org/10.1609/aaai.v35i16.17683} {{Meta-Curriculum
  Learning for Domain Adaptation in Neural Machine Translation}}.
\newblock In \emph{Proceedings of the AAAI Conference on Artificial
  Intelligence (AAAI)}, volume~35, pages 14310--14318.

\end{thebibliography}

\appendix

\section{Hyperparameter Setup}
We fix the decaying masking ratio schedule $\mathbf{r}= \{r_1, \dots, r_K\}$, that is employed in Algorithm~\ref{masking_alg}, through validation. For each architecture, we mention the maximum and minimum masking ratios in Table~\ref{table_hyperparams}, as well as the length $K$. We cycle through the schedules every $K$ epochs. We emphasize that the number of masking ratios $K$ determines the cycle length, i.e.~the masking ratios represent the curriculum probabilities that are cycled. In other words, the length of the masking ratio vector is equal to the cycle length. 

\begin{table}[t!]
    \centering
    \small{
    \begin{tabular}{|l| l| c|c|c|}
    \hline
         Model & Data Set & $r_1$& $r_K$& $K$ \\
         \hline
         \hline
         \multirow{5}{*}{BERT} 
             & Reuters & 0.15 & 0.09& 3\\
             & 20 News & 0.15 & 0.09& 3\\
             & SST2 &0.15& 0.09& 3\\
              & PAN19-P1 & 0.35 & 0.00 & 3\\
              & PAN19-P5 & 0.30 & 0.00 & 3\\
        \hline
        \multirow{5}{*}{RoBERTa}
             & Reuters & 0.15 & 0.09& 3\\
             & 20 News & 0.15 & 0.00 & 3\\
             & SST2 &0.15& 0.09& 3\\
              & PAN19-P1 & 0.35 & 0.00 & 3\\
              & PAN19-P5 & 0.30 & 0.00 & 3\\
             \hline
        \multirow{3}{*}{GPT-2}& SST2 &0.15& 0.09& 3\\
              & PAN19-P1 & 0.35 & 0.00 & 3\\
              & PAN19-P5 & 0.30 & 0.00 & 3\\
             \hline
         
    \end{tabular}
    }
    \vspace{-0.2cm}
    \caption{Masking ratios used in our experiments.}
    \label{table_hyperparams}
    \vspace{0.2cm}
\end{table}

\section{Ablation Studies}

\begin{table}[t]
  \centering
  \small{
  \begin{tabular}{|l|l|c|}
    \hline
    \multirow{2}{*}{Model} & \multirow{2}{*}{$r_1$-$r_K$} & SST2 \\
    \cline{3-3}
    & & Accuracy \\
    \hline
    \hline
    \multirow{4}{*}{BERT$_{\text{base}}$} 
      & 0.13-0.11 & $94.27_{\pm0.12}$ \\
      & 0.14-0.10 & $94.53_{\pm0.09}$ \\
      & 0.15-0.09 & $\mathbf{94.61}_{\pm0.08}$ \\
      & 0.16-0.08 & $94.58_{\pm0.06}$ \\
    \hline
    \multirow{4}{*}{RoBERTa$_{\text{base}}$} 
      & 0.13-0.11 & $94.82_{\pm0.18}$ \\
      & 0.14-0.10 & $94.95_{\pm0.21}$ \\
      & 0.15-0.09 & $95.04_{\pm0.18}$ \\
      & 0.16-0.08 & $\mathbf{95.06}_{\pm0.17}$ \\
    \hline
  \end{tabular}
  \vspace{-0.2cm}
  \caption{\label{ablation_r_sst2}Ablation study for intervals of masking ratios between $r_1$ and $r_K$ on SST2. The top score for each architecture is highlighted in bold.}
  }
\end{table}

\begin{table}[t]
  \centering
  \setlength\tabcolsep{0.28em}
  \small{
  \begin{tabular}{|c|l|c|c|c|}
    \hline
   {Data} &  \multirow{2}{*}{Model} &  \multirow{2}{*}{$r_1$-$r_K$} &  \multirow{2}{*}{Accuracy} &  \multirow{2}{*}{Macro F1} \\
   Set & & & & \\
    \hline
    \hline
    \multirow{8}{*}{\rotatebox[origin=c]{90}{PAN19-P1}} &
    \multirow{4}{*}{BERT$_{\text{base}}$} 
      & 0.35-0.00 & $77.32_{\pm9.33}$ & $\mathbf{60.60}_{\pm7.37}$ \\
     & & 0.37-0.02 & $76.99_{\pm8.83}$ & $58.89_{\pm7.64}$ \\
     & & 0.33-0.02 & $\mathbf{77.62}_{\pm8.21}$ & $59.94_{\pm7.23}$ \\
     & & 0.30-0.05 & $68.52_{\pm15.72}$ & $57.01_{\pm8.81}$ \\
    \cline{2-5}
    & \multirow{4}{*}{RoBERTa$_{\text{base}}$} 
      & 0.35-0.00 & $\mathbf{93.98}_{\pm0.70}$ & $\mathbf{83.78}_{\pm2.55}$ \\
     & & 0.37-0.02 & $93.23_{\pm1.36}$ & $82.83_{\pm2.66}$ \\
     & & 0.33-0.02 & $92.74_{\pm1.01}$ & $83.15_{\pm3.21}$ \\
     & & 0.30-0.05 & $91.23_{\pm1.06}$ & $81.51_{\pm2.32}$ \\
    \hline
        \multirow{8}{*}{\rotatebox[origin=c]{90}{PAN19-P5}} &
    \multirow{4}{*}{BERT$_{\text{base}}$} 
      & 0.30-0.00 & $\mathbf{69.94}_{\pm1.98}$ & $\mathbf{44.20}_{\pm2.67}$ \\
     & & 0.32-0.02 & $69.45_{\pm1.83}$ & $43.33_{\pm2.76}$ \\
     & & 0.28-0.02 & $68.92_{\pm2.12}$ & $44.02_{\pm2.34}$ \\
     & & 0.35-0.05 & $68.84_{\pm2.67}$ & $42.20_{\pm2.05}$ \\
    \cline{2-5}
    & \multirow{4}{*}{RoBERTa$_{\text{base}}$} 
      & 0.30-0.00 & $68.38_{\pm1.53}$ & $41.86_{\pm2.15}$ \\
     & & 0.32-0.02 & $\mathbf{69.84}_{\pm1.54}$ & $42.23_{\pm5.71}$ \\
     & & 0.28-0.02 & $68.72_{\pm1.89}$ & $41.33_{\pm2.48}$ \\
     & & 0.35-0.05 & $67.61_{\pm2.55}$ & $\mathbf{42.33}_{\pm6.03}$ \\
    \hline
  \end{tabular}
  }
  \vspace{-0.2cm}
  \caption{Ablation study for intervals of masking ratios between $r_1$ and $r_K$ on PAN19. The top scores for each architecture and metric are highlighted in bold.}
  \label{ablation_r_panp1}
  \vspace{0.2cm}
\end{table}

\begin{table*}[t]
  \centering
  \small{
  \begin{tabular}{|l|c|c|c|c|c|c|}
    \hline
    \multirow{2}{*}{Model} & \multirow{2}{*}{$K$} 
    & \multicolumn{1}{c|}{SST2} & \multicolumn{2}{c|}{PAN19-P1} & \multicolumn{2}{c|}{PAN19-P5} \\
    \cline{3-7}
    & & Accuracy & Accuracy & Macro F1 & Accuracy & Macro F1 \\
    \hline
    \hline
    \multirow{4}{*}{BERT$_{\text{base}}$} 
      & 2 & $93.97_{\pm0.11}$ & $66.68_{\pm14.06}$ & $55.52_{\pm5.95}$ & $66.28_{\pm1.13}$ & $42.64_{\pm1.51}$ \\
      & 3 & $\mathbf{94.61}_{\pm0.08}$ & $\mathbf{77.32}_{\pm9.33}$ & $\mathbf{60.60}_{\pm7.37}$ & $\mathbf{69.94}_{\pm1.98}$ & $\mathbf{44.20}_{\pm2.67}$ \\
      & 4 & $94.28_{\pm0.22}$ & $73.40_{\pm10.04}$ & $59.87_{\pm6.97}$ & $69.17_{\pm1.94}$ & $43.99_{\pm1.89}$ \\
      & 5 & $94.55_{\pm0.12}$ & $72.80_{\pm8.24}$ & $58.80_{\pm8.92}$ & $68.78_{\pm4.10}$ & $41.24_{\pm5.21}$ \\
    \hline
    \multirow{4}{*}{RoBERTa$_{\text{base}}$} 
      & 2 & $94.68_{\pm0.16}$ & $93.37_{\pm1.18}$ & $81.93_{\pm3.63}$ & $67.89_{\pm2.32}$ & $40.98_{\pm3.31}$ \\
      & 3 & $\mathbf{95.04}_{\pm0.18}$ & $\mathbf{93.98}_{\pm0.70}$ & $\mathbf{83.78}_{\pm2.55}$ & $68.38_{\pm1.53}$ & $41.86_{\pm2.15}$ \\
      & 4 & $94.91_{\pm0.21}$ & $92.85_{\pm1.20}$ & $83.41_{\pm2.30}$ & $\mathbf{71.30}_{\pm3.66}$ & $\mathbf{45.97}_{\pm3.15}$ \\
      & 5 & $95.02_{\pm0.16}$ & $92.62_{\pm1.39}$ & $82.38_{\pm1.56}$ & $68.93_{\pm2.14}$ & $41.50_{\pm1.89}$ \\
    \hline
  \end{tabular}
  }
  \vspace{-0.2cm}
  \caption{\label{tab_ablation_k}Ablation study for the hyperparameter $K$ on SST2 and PAN19. The top scores for each architecture and metric are highlighted in bold.}
\end{table*}

\noindent
\textbf{Ablating the masking ratios.}
In Tables \ref{ablation_r_sst2} and \ref{ablation_r_panp1}, we ablate the interval of masking ratios $r_1$-$r_K$, while maintaining the cycle length $K=3$. The reported results indicate that TIACBM maintains its performance gains for various intervals. This observation suggests that suboptimal hyperparameter choices can still lead to considerable improvements, attesting the robustness of TIACBM.

\noindent
\textbf{Ablating the cycle length.}
We perform additional experiments by ablating the cycle length $K$ between $2$ and $5$, while keeping the minimum and maximum masking ratios fixed. The corresponding results are shown in Table~\ref{tab_ablation_k}. While there are signs of sensitivity to the value of $K$, the results are generally above the conventional fine-tuning strategy. We conclude that $K$ should be carefully tuned for optimal results.

\begin{table*}[ht]
  \centering
  \setlength\tabcolsep{0.23em}
  \small{
  \begin{tabular}{|l|c|c|c|c|c|c|c|c|}
    \hline
      \multirow{2}{*}{Model} & \multicolumn{1}{c|}{Fine-Tuning}  & Reuters & 20 News & SST2 & \multicolumn{2}{|c|}{PAN19-P1} & \multicolumn{2}{c|}{PAN19-P5}\\
     \cline{3-9}
     & \multicolumn{1}{c|}{Strategy} & \multicolumn{2}{c|}{Micro F1} & Accuracy & Accuracy & Macro F1 & Accuracy & Macro F1\\
     \hline
    \hline 
    \multirow{2}{*}{BERT$_{\text{base}}$}& TICBM & $90.53_{\pm0.15}$ & $85.38_{\pm0.08}$ & $94.03_{\pm0.16}$  & $64.43_{\pm3.19} $ & $52.17_{\pm8.35}$ & $63.90_{\pm2.51} $ & $ 35.23_{\pm3.56}$\\
    & TIACBM & $\mathbf{91.20}_{\pm0.20}$ & $\mathbf{85.65}_{\pm0.10}$ & $\mathbf{94.61}_{\pm0.08}$ & $\mathbf{77.32}_{\pm9.33}$ & $\mathbf{60.60}_{\pm7.37}$ &$\mathbf{69.94}_{\pm1.98}$ & $\mathbf{44.20}_{\pm2.67}$ \\
    \hline
    
    \multirow{2}{*}{RoBERTa$_{\text{base}}$}& TICBM & $90.35_{\pm0.09}$ & $85.45_{\pm0.13}$ & $94.92_{\pm0.14}$& $90.93_{\pm1.22} $ & $ 78.83_{\pm0.98}$ &  $67.92_{\pm0.95} $ & $ 39.26_{\pm1.61}$ \\
    & TIACBM & $\mathbf{91.06}_{\pm0.19}$ & $\mathbf{85.93}_{\pm0.18}$& $\mathbf{95.04}_{\pm 0.18}$&$\mathbf{93.98}_{\pm0.70}$ & $\mathbf{83.78}_{\pm2.55}$ & $\mathbf{68.38}_{\pm1.53}$ & $\mathbf{41.86}_{\pm2.15}$\\
    \hline
  \end{tabular}
  }
  \vspace{-0.2cm}
  \caption{\label{curriulum_vs_anticurriculum}Comparison between curriculum (easy-to-hard) and anti-curriculum (hard-to-easy) approaches applied to BERT and RoBERTa. Both methods benefit from a cyclic schedule and task-specific information. The top score for each metric is highlighted in bold.}
\end{table*}

\begin{table*}[htb]
  \centering
  \setlength\tabcolsep{0.6em}
  \small{
  \begin{tabular}{|c|c|c|c|c|c|c|}
    \hline
    \multirow{2}{*}{Model} & \multirow{2}{*}{Fine-Tuning Strategy} 
    & SST2 & \multicolumn{2}{c|}{PAN19-P1} & \multicolumn{2}{c|}{PAN19-P5} \\
    \cline{3-7}
    & & Accuracy & Accuracy & Macro F1 & Accuracy & Macro F1 \\
    \hline
    \hline
    \multirow{2}{*}{GPT-2} 
    & TICBM 
    & $92.71_{\pm0.28}$ 
    & $83.62_{\pm3.73}$ & $70.86_{\pm5.36}$ 
    & $66.54_{\pm3.14}$ & $40.28_{\pm4.53}$ \\
    
    & TIACBM 
    & $\mathbf{92.96}_{\pm0.14}$ 
    & $\mathbf{85.90}_{\pm1.96}$ & $\mathbf{73.44}_{\pm3.81}$ 
    & $\mathbf{68.20}_{\pm2.25}$ & $\mathbf{42.90}_{\pm2.56}$ \\
    \hline
  \end{tabular}
  }
  \vspace{-0.2cm}
  \caption{Comparison between curriculum (easy-to-hard) and anti-curriculum (hard-to-easy) approaches applied to GPT-2. The top score for each metric is highlighted in bold.}
  \label{tab:gpt2_curriulum_vs_anticurriculum}
\end{table*}

\begin{table*}[htb]
\centering
\setlength\tabcolsep{0.4em}
\small{
\begin{tabular}{|l|l|c|c|c|c|}
\hline
\multirow{2}{*}{Model} & \multirow{2}{*}{Method} & \multicolumn{2}{c|}{PAN19-P1} & \multicolumn{2}{c|}{PAN19-P5} \\
\cline{3-6}
& & Accuracy & Macro F1 & Accuracy & Macro F1\\
\hline
\hline
\multirow{3}{*}{BERT$_{\text{base}}$} 

& Conventional & $69.76_{\pm15.11} $ & $58.24_{\pm6.06}$&$66.10_{\pm1.56} $ & $36.10_{\pm4.22}$\\

  & Sentiment Heuristic & $76.62_{\pm8.05}$ & $59.44_{\pm3.48}$ & $\mathbf{70.20}_{\pm0.36} $ & $ \mathbf{45.33}_{\pm2.90}$ \\
  & Authorship Heuristic (Original) & $\mathbf{77.32}_{\pm9.33} $ & $ \mathbf{60.60}_{\pm7.37}$ & $69.94_{\pm1.98} $ & $ 44.20_{\pm2.67}$ \\
\hline
\multirow{3}{*}{RoBERTa$_{\text{base}}$} 

& Conventional &$89.20_{\pm3.01} $ & $76.92_{\pm4.64}$&$67.42_{\pm2.90} $ & $38.30_{\pm4.32}$\\

  & Sentiment Heuristic & $92.06_{\pm1.52} $ & $ 81.72_{\pm1.81}$ & $\mathbf{69.68}_{\pm5.50} $ & $ \mathbf{46.40}_{\pm5.71}$ \\
  & Authorship Heuristic (Original) & $\mathbf{93.98}_{\pm0.70} $ & $ \mathbf{83.78}_{\pm2.55}$ & $68.38_{\pm1.53} $ & $ 41.86_{\pm2.15}$ \\
\hline
\end{tabular}
}\vspace{-0.2cm}
\caption{\label{task_transfer}Results obtained by transferring our sentiment heuristic (originally applied on SST2) to authorship attribution (PAN19-P1 and PAN19-P5), for both BERT and RoBERTa. The task-specific heuristics used by TIACBM can be applied across tasks. The top scores for each architecture and metric are highlighted in bold.}
\end{table*}

\begin{table*}[htb]
  \centering
  \small{
  \begin{tabular}{|l|l|c|c|c|c|c|}
    \hline
    \multirow{2}{*}{{Model}} & \multirow{2}{*}{{Heuristic}} 
    & SST2 & \multicolumn{2}{c|}{PAN19-P1} & \multicolumn{2}{c|}{PAN19-P5} \\
    \cline{3-7}
    & & Accuracy & Accuracy & Macro F1 & Accuracy & Macro F1 \\
    \hline
    \hline
    \multirow{3}{*}{BERT$_{\text{base}}$} 
    & Conventional & $93.38_{\pm0.14}$ & $69.76_{\pm15.11} $ & $58.24_{\pm6.06}$&$66.10_{\pm1.56} $ & $36.10_{\pm4.22}$\\

      & Generic & $94.07_{\pm0.18}$ & $69.78_{\pm16.83} $ & $ 58.24_{\pm11.61}$ & $\mathbf{71.06}_{\pm2.46} $ & $ \mathbf{47.14}_{\pm3.75}$ \\
      & Task-specific & $\mathbf{94.61}_{\pm0.08}$ & $\mathbf{77.32}_{\pm9.33} $ & $ \mathbf{60.60}_{\pm7.37}$ & $69.94_{\pm1.98} $ & $ 44.20_{\pm2.67}$ \\
    \hline
    \multirow{3}{*}{RoBERTa$_{\text{base}}$} 
    & Conventional & $94.56_{\pm0.09}$ &$89.20_{\pm3.01} $ & $76.92_{\pm4.64}$&$67.42_{\pm2.90} $ & $38.30_{\pm4.32}$\\

      & Generic & $94.83_{\pm0.16}$ & $91.73_{\pm0.56} $ & $ 80.75_{\pm1.06}$ & $\mathbf{71.96}_{\pm0.61} $ & $ \mathbf{42.96}_{\pm2.97}$ \\
      & Task-specific & $\mathbf{95.04}_{\pm0.18}$ & $\mathbf{93.98}_{\pm0.70} $ & $ \mathbf{83.78}_{\pm2.55}$ & $68.38_{\pm1.53} $ & $ 41.86_{\pm2.15}$ \\
    \hline
  \end{tabular}
  \vspace{-0.2cm}
  }
  \caption{\label{general_task_method}Results with the generic (task-agnostic) attention-based heuristic versus task-specific heuristics. The task-agnostic heuristic can be effective when task-specific information is not available. The top scores for each architecture and metric are highlighted in bold.}
\end{table*}

\section{Curriculum versus Anti-Curriculum}

In Tables~\ref{curriulum_vs_anticurriculum} and \ref{tab:gpt2_curriulum_vs_anticurriculum}, we compare our anti-curriculum method (TIACBM) with a reversed masking ratio schedule, which implements an easy-to-hard curriculum learning. Both approaches benefit from a cyclic schedule and task-specific information. The anti-curriculum approach consistently outperforms its counterpart, validating our choice based on hard-to-easy curriculum.

\section{TIACBM without Linguistic Priors}

\noindent
\textbf{Transferring task heuristics.}
To show that the domain-specific heuristics can generalize across domains, we conduct experiments with the sentiment heuristic (originally applied on SST2) on authorship attribution. In Table~\ref{task_transfer}, we show the corresponding results with BERT and RoBERTa. We observe that the sentiment heuristic is either close to the authorship heuristic, or even surpasses it in performance. This can be seen in the case of PAN19-P5, where the sentiment heuristic capitalizes on the data set design (fandom creative writing) and benefits from the numerous subjective words. Overall, the empirical evaluation indicates that the proposed task-specific heuristics can generalize across tasks. Furthermore, we observe that the sentiment heuristic consistently outperforms the conventional fine-tuning regime, suggesting that TIACBM can bring performance gains even when its heuristic is not aligned with the downstream task.

\noindent
\textbf{Generic versus task-specific heuristics.}
We next employ a generic task-agnostic heuristic, solely based on the attention scores of the fine-tuned model. The generic heuristic masks a number of tokens at each epoch, where the probability of masking a token is proportional to its average weight. We only utilize the attention weights from the first epoch, storing and loading them in later epochs. As shown in Table~\ref{general_task_method}, the generic attention-based approach provides competitive results with our task-specific approaches, even surpassing them in a few cases. Task-specific heuristics provide better overall results for specific tasks, while the generic attention-based heuristic excels in some areas and lacks in others. The generic approach represents an alternative when there are no task-specific priors that can be leveraged.

\noindent
\textbf{Overall assessment.}
In summary, the experiments presented in Table \ref{task_transfer} and \ref{general_task_method} show that our text-based anti-curriculum by masking does not necessarily depend on prior task-specific information. 

\section{Computational Resources}
The experiments are carried out on a machine with 64GB of RAM, an AMD Ryzen 7 7800X3D CPU, and an Nvidia GeForce RTX 4090 GPU. Our most expensive experiments are performed on 20 Newsgroups, with 112 \texttt{mins} (3 \texttt{mins} per epoch) for TIACBM, and about 180 \texttt{mins} for decaying runs, due to the need of masking at every epoch. The masking step takes between 2-3 \texttt{mins}. In the case of SST2, the experiments require 6 \texttt{mins} per epoch, while the masking step requires 0.5 \texttt{mins}. The timetable for PAN19 varies from 0.5 \texttt{mins} per epoch to 1 \texttt{min} per epoch, regardless of the approach, while the masking takes roughly 3-4 \texttt{secs}. Finally, the masking step on Reuters-21578 takes between 1-3 \texttt{mins}, depending if non-content words need to be discarded or not. An average epoch takes 1.5 \texttt{mins}.

\end{document}